\pdfoutput=1

\documentclass[11pt]{article}

\usepackage[]{acl}

\usepackage{times}
\usepackage{latexsym}
\usepackage{url}
\usepackage{latexsym}
\usepackage{multicol}
\usepackage{multirow}
  \usepackage{graphicx}  
  \usepackage{epstopdf}
  \usepackage{amssymb}
   \usepackage{color}
   \usepackage{amsmath} 
  \usepackage{xcolor}
  \usepackage{longtable}
\usepackage[T1]{fontenc}

\usepackage[utf8]{inputenc}
\usepackage{bbm}
\usepackage{microtype}
\usepackage{hyperref}[colorlinks,linkcolor=blue]
\title{Mere Contrastive Learning for Cross-Domain Sentiment Analysis}
\author{
    Yun Luo \textsuperscript{\rm1,2},
    Fang Guo \textsuperscript{\rm2},
    Zihan Liu \textsuperscript{\rm2},
    Yue Zhang \textsuperscript{\rm2,3} 
    \\
    \textsuperscript{1} School of Computer Science And Technology, Zhejiang University, Hangzhou 310024, P.R. China. \\
    \textsuperscript{2} School of Engineering, Westlake University, Hangzhou, China. \\
    \textsuperscript{3} Institute of Advanced Technology, Westlake Institute for Advanced Study, Hangzhou, China.  \\
    \texttt{\{luoyun, guofang, liuzihan, zhangyue\}@westlake.edu.cn}\\
}

\begin{document}
\maketitle
\begin{abstract}
Cross-domain sentiment analysis aims to predict the sentiment of texts in the target domain using the model trained on the source domain to cope with the scarcity of labeled data. Previous studies are mostly cross-entropy-based methods for the task, which suffer from instability and poor generalization.  In this paper, we explore contrastive learning on the cross-domain sentiment analysis task. We propose a modified contrastive objective with in-batch negative samples so that the sentence representations from the same class will be pushed close while those from the different classes become further apart in the latent space. Experiments on two widely used datasets show that our model can achieve state-of-the-art performance in both cross-domain and multi-domain sentiment analysis tasks. Meanwhile, visualizations demonstrate the effectiveness of transferring knowledge learned in the source domain to the target domain and the adversarial test verifies the robustness of our model.
\end{abstract}

\section{Introduction}
Sentiment classification \cite{liu2012sentiment} has been widely studied by both industry and academia \cite{blitzer2007biographies,li2013active,yu2016learning}. For example, the sentiment is positive towards the text \textit{`The book is exactly as pictured/described. Cute design and good quality'}. Early methods rely on labeled data to train models on a specific domain (e.g. DVD reviews, book reviews, and so on), which are labor-intensive and time-consuming \cite{socher2013recursive}. To address this issue, cross-domain sentiment analysis attracts increasing attention.

\begin{figure}[t]
    \centering
    \includegraphics[width = 0.75\hsize]{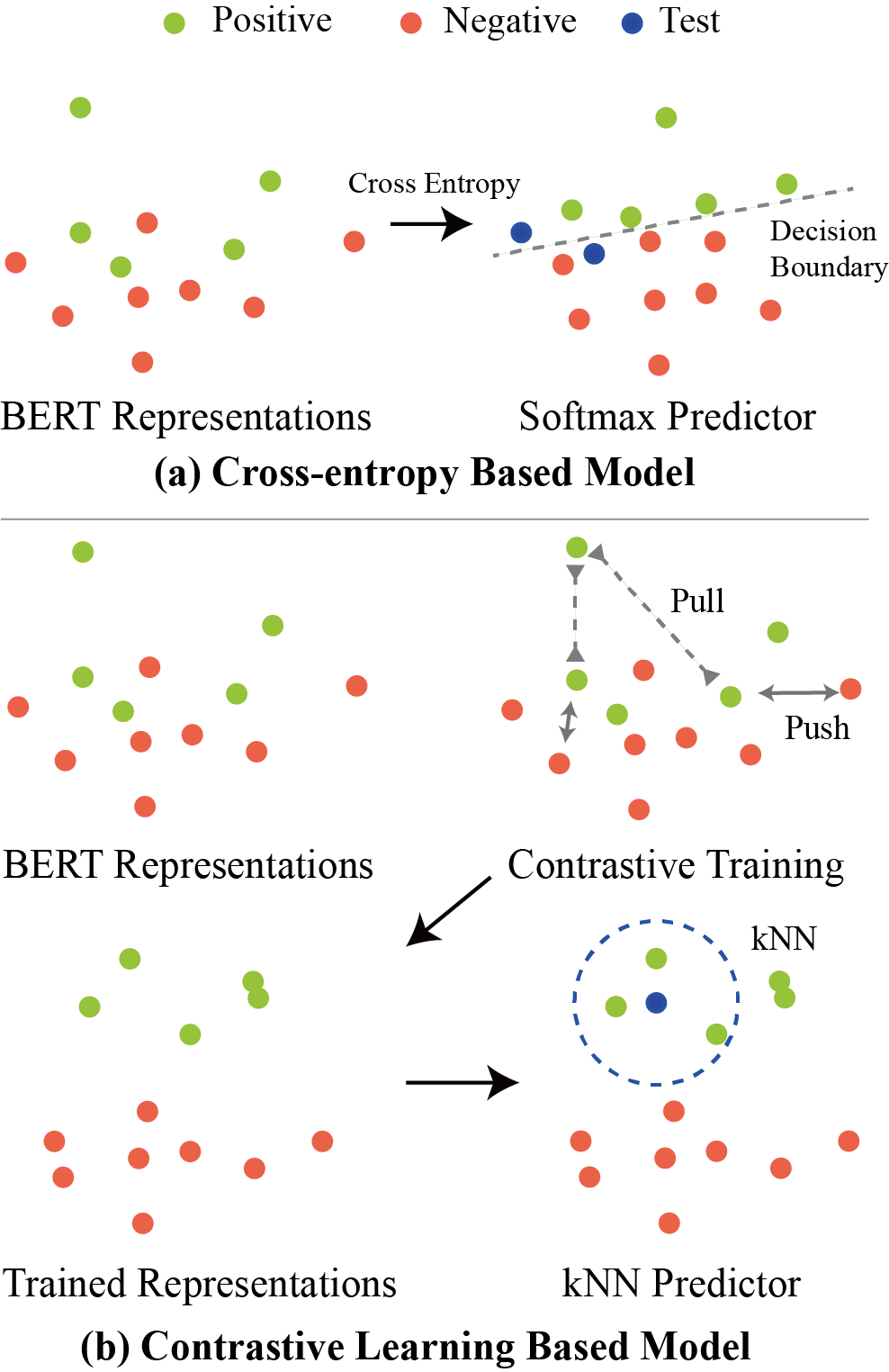}
    \caption{The architectures for cross-entropy-based model and the contrastive- learning-based model.}
    \label{fig1}
\end{figure}

Various neural models have been proposed for cross-domain sentiment analysis in recent years  \cite{blitzer2007biographies,li2013active,yu2016learning,zhang2019interactive,zhou2020sentix}. Most methods focus on making the model unable to distinguish the data from which domain by adversarial training, in order to transfer knowledge from source domains to target domains \cite{du2020adversarial,liu2017adversarial,qu2019adversarial} and some attempt to learn domain-specific knowledge \cite{zhou2020sentix,Liu2018,wang2019learning}. Pre-trained language models \cite{devlin2018bert,radford2019language,lewis-etal-2020-bart} have achieved stronger performance compared with previous random initialized models such as LSTM (Long Short-term Memory) in cross-domain tasks.  The state-of-the-art models on cross-domain sentiment analysis, such as BERT-DAAT \cite{du2020adversarial}, use unlabeled data to continually train the pre-trained model BERT to transfer knowledge besides adversarial training. 

In the representation aspect for the cross-domain sentiment analysis, there are two key requirements for the representations of sentences: (1)  sentence representations in the same domain with the different/the same sentiments should be far from/close to each other; (2)  sentence representations of different domains with the same labels should be close.  Existing methods are mostly softmax-based method by optimizing cross-entropy loss to achieve the requirements (illustrate in Figure \ref{fig1} (a)), which suffers from instability across different runs \cite{zhang2020revisiting,Dodge}, poor generalization performance \cite{liu2016large,cao2019learning}, reduction of prediction diversity \cite{cui2020towards}, lack of robustness to noisy labels \cite{zhang2018generalized,sukhbaatar2015training}, or adversarial examples \cite{elsayed2018large,nar2019cross}, especially when supervised data are limited in the cross-domain settings.

To address the above shortcomings, we explore the effectiveness of contrastive learning on the task.
Contrastive learning is a similarity-based training strategy, which aims to push the representations from the same class close and those from the different class further apart \cite{Chen2020,gao2021simcse,neelakantan2022text,gao2021simcse}.   Contrastive learning has been shown effective in solving the problem of anisotropy \cite{Gao2019}, and it has good generalization and robustness \cite{li2021knn,gao2021simcse,gunel2020supervised,khosla2020supervised}. Previous work relies mostly on pre-training for representations \cite{Chen2020,neelakantan2022text} or multi-task training for semantic textual similarity \cite{gao2021simcse},  classification \cite{li2021knn,gunel2020supervised} and so on, but little work uses mere contrastive learning for supervised tasks.  Intuitively, the optimization of contrastive learning is effective in satisfying the requirements of cross-domain sentiment analysis.




We explore \textbf{CO}ntrastive learning on \textbf{BE}RT (COBE) by a modified contrastive loss function with the in-batch negative method on cross-domain sentiment analysis tasks. In the mini-batch, the samples with the same labels are treated as positive pairs, and those with different labels are treated as negative pairs. As shown in Figure \ref{fig1}, the optimization procedure  aims to tighten the
cluster of samples with the same labels, and push away samples with different labels. After training, the representations of training data and their labels are saved offline as a knowledge base for classification. When evaluating the model, a kNN (k-Nearest Neighbors) predictor is used to predict the sentiment of test data, i.e. we search for the $k$ data with the largest cosine similarity in the knowledge base and vote for the final prediction using their labels.

Experiments on two widely used datasets (the cross-domain Amazon dataset \cite{blitzer2007biographies} and  FDU-MTL \cite{liu2017adversarial}) show that our model can achieve the state-of-the-art performance in both the cross-domain setting and the multi-domain setting sentiment classification. Visualizations also demonstrate the effectiveness of transferring knowledge learned in the source domain to the target domain. To our knowledge, we are the first to show that contrastive learning outperforms cross-entropy-based models on cross-domain sentiment analysis for both performance and robustness. The code has been released in \href{https://github.com/LuoXiaoHeics/COBE}{https://github.com/LuoXiaoHeics/COBE}. 

\begin{figure*}
    \centering
    \includegraphics[width =0.9\hsize]{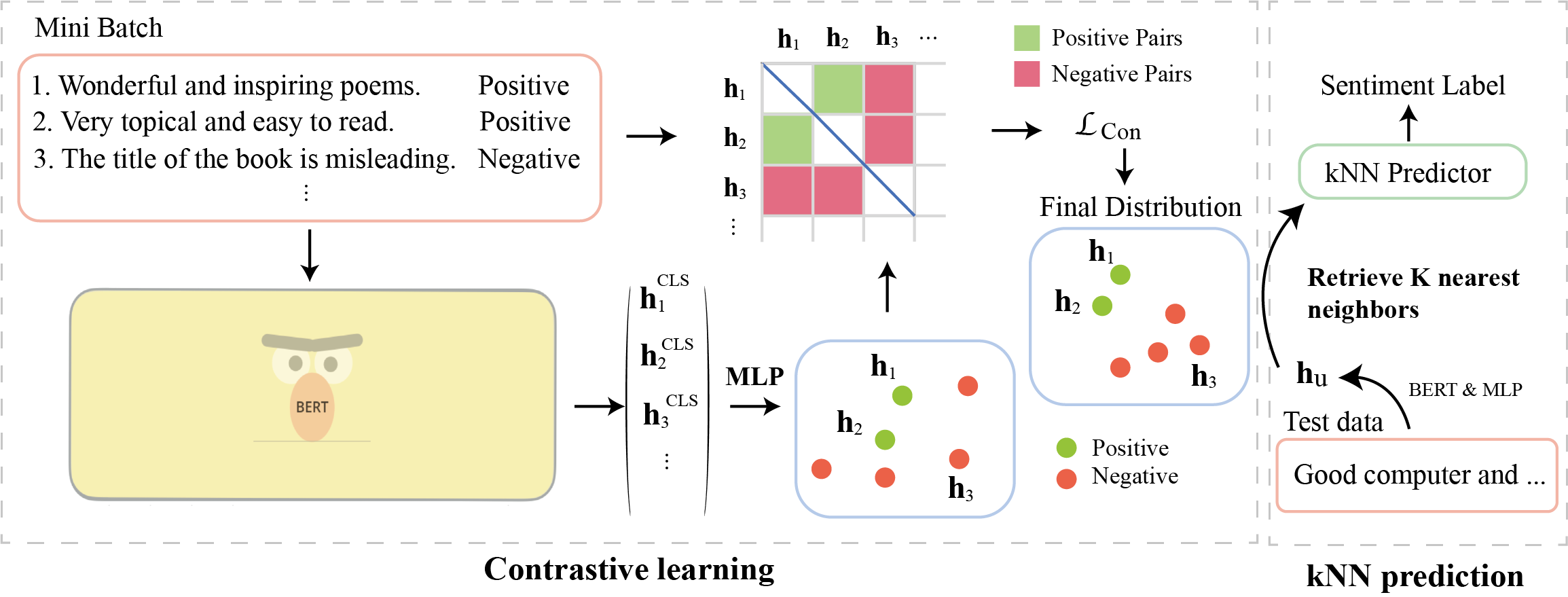}
    \caption{The framework of our contrastive learning for cross-domain sentiment analysis.}
    \label{model}
\end{figure*}

\section{Related Work}
\textbf{Cross-domain sentiment analysis.}  Due to the heavy cost of obtaining large quantities of labeled data for each domain, many approaches have been proposed for cross-domain sentiment analysis \cite{blitzer2007biographies,li2013active,yu2016learning,zhang2019interactive,zhou2020sentix}.  \citet{ziser2018pivot} and \citet{li2018hierarchical} propose to capture the pivots that are useful for both source domains and target domains. \citet{ganin2016domain} propose to use adversarial training with a domain discriminator to learn domain-invariant information, which is one type of solutions for the cross-domain sentiment analysis task \cite{du2020adversarial,liu2017adversarial,qu2019adversarial,zhou2020sentix}.  These adversarial training methods try to confuse the model unable to classify the data from which domain, transferring the knowledge from source domains to target domains.  Besides, \citet{Liu2018} and \citet{cai2019multi} attempt to learn domain-specific information for the different sentiment expressions on different domains. However, these studies rely on minimizing the cross-entropy loss, resulting in the issue of unstable fine-tuning and poor generalization \cite{gunel2020supervised,li2021knn,zhang2020revisiting,Dodge}. 

\textbf{Contrastive Learning.} Contrastive learning has been widely used in unsupervised learning \cite{Chen2020,jing2021understanding,wang2020understanding,khosla2020supervised,gao2021simcse,neelakantan2022text}. 
\citet{radford2019language} propose to use contrastive learning to learn the representations of both text and images through raw data in unsupervised method, which achieves strong performance on zero-shot task. \citet{neelakantan2022text} propose to use contrastive learning to obtain sentence and code representations and achieve strong performance on downstream tasks such as sentence classification and text search. \citet{wang2020understanding} further identify the key properties for contrastive learning as (1) alignment (closeness) of features from positive pairs and (2) uniformity of induced distribution of representations. \citet{gao2021simcse} uses contrastive learning to learn the sentence representations and theoretically prove that contrastive learning can solve the anisotropy problem (the learned embeddings occupy a narrow cone in the vector space), which limits the expressiveness of representations. It also achieves better results for the semantic textual similarity  task using a supervised dataset of 
natural language inference. Our model differs from the above studies in that we consider contrastive learning in supervised tasks, which uses golden labels to obtain positive/negative pairs for training.

Recently, some studies attempt to incorporate contrastive learning into cross-entropy-based methods by adding InfoNCE loss \cite{gunel2020supervised,li2021knn}, which aims to solve the shortcomings of cross-entropy loss. \citet{gunel2020supervised} propose a new SCL loss based on InfoNCE loss to boost the stability and robustness of fine-tuning pre-trained language models. Subsequently, \citet{li2021knn} attempt to incorporate kNN predictors to enhance the generalization of prediction in few-shot tasks, using both cross-entropy loss and SCL loss. The above work is similar to ours in making use of contrastive loss for classification. However, the difference is that we do not use a standard cross-entropy loss, but  rely solely on vector space similarity losses for achieving cross-domain classification. To our knowledge, we are the first to conduct sentiment classification without using a cross-entropy loss in natural language processing.

\section{Method}
Formally, the training data consists of $\{(S_i,Y_i)\}_{i=1}^N$, where $S_i = [s_1,s_2,...,s_l]$ is a set of review text, and $Y_i\in \{0,1\}$ are the corresponding sentiment labels. The model framework is shown in Figure \ref{model}. We first introduce the  prediction of sentiment labels using representations based on kNN (Section 3.1), and then describe the training objective to obtain effective representations using  contrastive learning (Section 3.2). For comparison, we also describe the standard cross-entropy baseline, named BERT-CE, and a version that  adopts adversarial training, named BERT-adv.
\subsection{Model}
We concatenate the review text $S_i$ with special tokens $[CLS]$ and $[SEP]$ as our model input $X_i = \ [CLS]\ S_i \ [SEP]$, which is fed into BERT model to obtain the hidden states. The hidden state of $[CLS]$ from the last layer of BERT is considered as the representation of the input sequence:
\begin{equation}
    \textbf{h}_i^{CLS} = BERT(X_i)[CLS]
\end{equation}

\textbf{BERT-CE and BERT-adv baselines}: After obtaining the sentence representation $\textbf{h}_i^{CLS}$ of input $X_i$, an MLP (Multi Layer Perceptron) layer project it to the label space and a softmax layer is adopted to  calculate the probability distribution on the labels: 
\begin{equation}
    p_{ce} = Softmax(MLP(\textbf{h}_i^{CLS})).
\end{equation}

The label with the largest probability is adopted as the prediction result.

\textbf{COBE}: Our model uses the same representation of Eq(1), but adopts a kNN predictor to classify the labels. An MLP layer is then adopted for the dimension reduction:
\begin{equation}
    \textbf{h}_i = MLP(\textbf{h}_i^{CLS}).
\end{equation}

To predict the sentiment label of a review text $S_u$, we calculate the cosine similarity of the sentence representation $\textbf{h}_u$ with the sentence representations of the training data:
\begin{equation}
sim(\textbf{h}_u, \textbf{h}_i) = \frac{\textbf{h}_u\cdot \textbf{h}_i}{||\textbf{h}_u|| \cdot ||\textbf{h}_i||},
\end{equation}
where $\textbf{h}_i$ is the sentence representation of training data $X_i$. 

We retrieve the $k$ training data whose cosine similarity with $\textbf{h}_u$ are the largest. We denote the $k$ nearest neighbors as ${(\textbf{h}_i,Y_i)\in \mathcal{K}_u}$. The retrieved set is converted to a probability distribution over the labels by applying a softmax with temperature $T$ to the similarity. Using the temperature $T>1$ can flatten the distribution, and prevent over-fitting to the most similar searches \cite{khandelwal2020nearest}. The probability distribution on the labels can be expressed as follows:
\begin{equation}
p_{k}(Y_u') \propto \sum_{(\textbf{h}_i,Y_i)\in \mathcal{K}_u} \mathbbm{1}_{Y_u'=Y_i} \cdot exp(\frac{sim(\textbf{h}_u,\textbf{h}_i)}{T} ).
\end{equation}

The label with the largest probability is regarded as the prediction result.

\subsection{Training Objective}














\textbf{BERT-CE baseline}: For the cross-entropy-based model, multi-label cross-entropy loss is adopted to optimize the model, which is formulated as follows:
\begin{equation}
    \mathcal{L}_{cls} = -\frac{1}{M}\sum_{(X_i,Y_i)}Y_ilog\ p_{ce}(Y_i),
\end{equation}

\textbf{BERT-adv baseline}:
Besides the cross-entropy loss, BERT-adv adds a domain discriminator \cite{du2020adversarial,ganin2016domain} to the standard model and  adopts adversarial training to transfer knowledge from source domains to target domains. 

Given the sentence and its domain label $(X_i,D_i)$, the representation $\textbf{h}_i^{CLS}$ obtained in Eq(1) goes through an additional gradient reversal layer (GRL) \cite{ganin2016domain}, which can be denoted as a `pseudo-function' $D_{\lambda}(x)$. The GRL reverses the gradient by applying a negative scalar $\lambda$. The forward- and backward- behaviors 
can be described:
\begin{equation}
    D_\lambda(x) = x, \ \frac{\partial D_\lambda(x)}{\partial x}  = -\lambda I,
\end{equation}
where $\lambda$ is a hyper-paramter and $I$ is the gradients calculated on  $\textbf{h}_i^{CLS}$ (but it is multiplied with $-\lambda$ to back-propagate). Then a linear layer project $\textbf{h}_i^{CLS}$ to the label space and a softmax layer is adopted to calculate the distribution on domain labels:
\begin{equation}
   p_d = Softmax(W_d\textbf{h}_i^{CLS}+b_d),
\end{equation}
where $W_d$ and $b_d$ are the learnable parameters. The training target is to minimize the cross-entropy for all data from the source and target domains (note that the data from target domains are unlabeled on sentiment) in order to make the model unable to predict the domain labels:
\begin{equation}
   \mathcal{L}_{dom} = -\frac{1}{M}\sum_{(X_i,D_i)}D_ilog\ p_d(D_i).
\end{equation}
For BERT-adv, the training loss of sentiment classification (Eq.7) and domain classification (Eq.10) are jointly optimized:
\begin{equation}
   \mathcal{L}_{adv} = \mathcal{L}_{cls} +  \mathcal{L}_{dom} 
\end{equation}

\begin{table*}[]\small
\centering
\begin{tabular}{l|p{0.035\textwidth}p{0.035\textwidth}p{0.035\textwidth}p{0.035\textwidth}p{0.035\textwidth}p{0.035\textwidth}p{0.035\textwidth}p{0.035\textwidth}p{0.035\textwidth}p{0.035\textwidth}p{0.035\textwidth}p{0.04\textwidth}|c}
\hline

S $\to$ T & B$\to$D & B$\to$E & B$\to$K & D$\to$B & D$\to$E & D$\to$K   & E$\to$B & E$\to$D & E$\to$K & K$\to$B & K$\to$D & K$\to$E & Avg   \\
\hline
\hline
DANN                        & 82.30             & 77.60             & 76.10             & 81.70             & 79.70             & 77.35 & 78.55             & 79.70             & 83.95             & 79.25             & 80.45             & 86.65             & 80.29 \\
PBLM                        & 84.20             & 77.60             & 82.50             & 82.50             & 79.60             & 83.20 & 71.40             & 75.00             & 87.80             & 74.20             & 79.80             & 87.10             & 80.40 \\
HATN                        & 86.10             & 85.70             & 85.20             & 86.30             & 85.60             & 86.20 & 81.00             & 84.00             & 87.90             & 83.30             & 84.50             & 87.00             & 85.10 \\
ACAN                        & 83.45             & 81.20             & 83.05             & 82.35             & 82.80             & 78.60 & 79.75             & 81.75             & 83.35             & 80.80             & 82.10             & 86.60             & 82.15 \\
IATN                        & 86.80             & 86.50             & 85.90             & 87.00             & 86.90             & 85.80 & 81.80             & 84.10             & 88.70             & 84.70             & 84.10             & 87.60             & 85.90 \\
\hline
BERT-CE                        & 88.96 &86.15 &89.05 &89.40&86.55&87.53 &86.50 &87.95 &91.60& 87.55 &87.95&90.45 &88.25\\
BERT-CE$^*$                   & 55.40             & 56.55             & 54.05             & 55.10             & 57.25             & 53.75 & 55.50             & 56.00             & 55.55             & 52.30             & 52.75             & 54.15             & 54.86 \\
BERT-adv &89.70&87.30&89.55&89.55&86.05&87.69&87.15&86.05&91.91&87.65&87.72&86.05&88.56\\ 
DAAT                   & 89.70             & 89.57             & 90.75             & 90.86             & 89.30             & 87.53 & \textbf{88.91}             & \textbf{90.13}             & 93.18             & 87.98             & \textbf{88.81}             & 91.72             & 90.12
\\
COBE$^*$& 82.17&83.65&83.12&79.82&78.87&82.58&75.95&79.53&86.10&78.55&76.95&85.17&80.95\\
COBE (proposed)&\textbf{90.05}&\textbf{90.45}&\textbf{92.90}&\textbf{90.98}&\textbf{90.67}&\textbf{92.00}&87.90&87.87&\textbf{93.33}&\textbf{88.38}&87.43&\textbf{92.58}&\textbf{90.39}\\

\hline
\end{tabular}
\caption{Results on the cross-domain Amazon dataset.  BERT-CE$^*$ and COBE$^*$ refer to the models fixing the parameters of BERT, and only tuning the parameters of MLP layer. (B for the Books domain, D for the DVD domain, E for the Electronics domain, and K for the Kitchen domain, respectively.)} 
\label{table2}
\end{table*}

\textbf{COBE}:  The baselines adopt  $\mathcal{L}_{cls}$ to tighten the representations of the same/different labels close (apart), and adopt $\mathcal{L}_{dom}$ to mix up the representations of different domains with the same label. However, COBE uses a single training objective of contrastive learning to achieve the both goals. We apply in-batch negatives \cite{yih2011learning,sohn2016improved} to learn sentence representations through contrastive learning, which has been widely used in unsupervised representation learning \cite{radford2021learning,jia2021scaling}. For each example in the mini-batch of ${M}$ samples,  we treat the other samples with different golden labels as negative pairs, and the samples with the same golden labels as positive pairs. For example in Figure \ref{model}, the sentence pair (1,2) is positive pairs, and the sentence pairs (1,3) and (2,3) are negative pairs.  For each review $X_i$  we denote $N_i^+$ as the set of reviews with the same label of $X_i$ in the mini-batch. Then the contrastive loss function can be defined as follows:
\begin{equation}
    \mathcal{L}_{Con} = \sum_i^M -\frac{1}{M}log \frac{\sum_{k\in N_i^+} exp(sim(\textbf{h}_i, \textbf{h}_k)/\tau)}{\sum_{i\neq j}^M exp(sim(\textbf{h}_i,\textbf{h}_j)/\tau)}
\end{equation}
where $\tau$ is a temperature hyper-parameter. 
The loss function can alleviate the negative effect of the situations where there is no positive pairs for any training instance in the batch.

The usage of in-batch negatives enables re-use of computation both in the forward and backward pass making training highly efficient.

\section{Experiments}
 We conduct experiments on both the cross-domain settings (train models on source domains and test on another one) and the multi-domain settings (train and test models on the same domains). To verify the effectiveness of our model --  \textbf{CO}ntrastive learning on \textbf{BE}RT (COBE), we also visualize the representations (Section 4.3) and carry out further analysis such as model robustness (Section 4.4). 

\subsection{Settings}
\textbf{Datasets.} We test our contrastive learning method on two widely used datasets, the cross-domain Amazon dataset, and the FDU-MTL dataset. The cross-domain Amazon dataset \cite{blitzer2007biographies} contains 4 domains: Books (B), DVD (D), Electronics (E) and Kitchen (K). Each domain contains 2000 Amazon review samples. Following the setting of previous work \cite{ganin2016domain,ziser2018pivot,du2020adversarial}, we test the model on 12 tasks. The model is trained on the source domain data and tested on the target domain data.

Furthermore, we also evaluate our model on FDU-MTL, which is an Amazon reviews dataset with data on 16 domains \cite{liu2017adversarial}. The training set, development set, and test set are split in the original dataset, (the statistics are shown in Appendix A). We carry out experiments on the multi-domain setting, (i.e. train the model on the whole 16 domains, and evaluate the model on the test on the whole 16 domains), and on the 15-1 cross-domain setting (i.e. train the model on 15 domains, and test the model on the  1 domain left).


\textbf{Baselines.}
For the cross-domain Amazon dataset, we compare our model with several strong baselines in cross-domain sentiment analysis: DANN \cite{ganin2016domain}, PBLM \cite{ziser2018pivot}, HATN \cite{Li_Wei_Zhang_Yang_2018}, IATN \cite{qu2019adversarial},  DAAT \cite{du2020adversarial}, BERT-CE and BERT-CE$^*$ ($^*$ for fixing the BERT parameters). We adopt the results of baselines reported in \citet{zhou-etal-2020-sentix} and \citet{du2020adversarial}.   
We also adopt BERT-adv as our baselines introduced in Section 3.2.

On FDU-MTL, we compare our model  with ASP \cite{liu2017adversarial},  DSR-at \cite{zheng2018same}, DAEA and DAEA-B (DAEA-BERT) \cite{cai2019multi}.  The DAEA-B is regarded as the state-of-the-art model on FDU-MTL (excluding the model SentiX \cite{zhou-etal-2020-sentix}, which uses a large corpus (about 241 million reviews) to continually train BERT for sentiment tasks). Note that, it is unfair that previous studies do not adopt the BERT-CE model for multi-domain experiments for comparison. In this study, we also consider BERT-CE, and BERT-CE$^*$ on the multi-domain setting as baselines. For the multi-domain task, the objective of adversarial training is redundant, thus we mainly compare COBE with baselines BERT-CE.

\begin{table*}[!htbp] \small
\centering

\begin{tabular}{l|ccccccccc}   
\hline  
Domain &ASP&
 DA & DSA & DAEA &DAEA-B  & BERT-CE$^*$ &BERT-CE& COBE$^*$ &COBE \\
\hline  
\hline
Books&84.00&88.50&89.10&89.00&N/A&81.33&\textbf{{90.67}} &85.17&{90.17}\\
Electronics&86.80 &89.00&87.90&91.80&N/A&82.17&91.92 &82.92&\textbf{93.58}\\
DVD &85.50&88.00&88.10&88.30&N/A&78.83&89.00&79.42&	\textbf{89.67}\\
Kitchen &86.20&89.00&85.90&90.30&N/A&79.92&91.17 &81.33&\textbf{91.50}\\
Apparel &87.00 &88.80&87.80&89.00&N/A&83.33&92.08&87.25&	\textbf{92.33}\\
Camera &89.20&91.80&90.00&92.00&N/A&81.83&93.25&87.50&\textbf{93.58}
\\
Health &88.20 &90.30&92.90&89.80&N/A&81.25&93.33	&85.00&\textbf{93.92}
\\
Music&82.50 &85.00&84.10&88.00&N/A&79.42&88.92&80.33& \textbf{90.33}
\\
Toys&88.00 &89.50&85.90&91.80&N/A&78.25&92.41&83.75&	\textbf{93.42}
\\
Video &84.50&89.50&90.30&\textbf{92.30}&N/A&78.17&90.33&83.67&	89.91
\\
Baby &88.20&90.50&91.70&92.30&N/A&82.33&93.00&84.42	&\textbf{93.92}
\\
Magazines &92.20&92.00&92.10&\textbf{96.50}&N/A&83.41&93.75&89.67&	94.08
\\
Software &87.20&90.80&87.00&92.80&N/A&83.42&92.42&85.33&	\textbf{93.42}
\\
Sports &85.70&89.80&85.80&90.80&N/A&78.50	&91.50&84.50&\textbf{92.83}
\\
IMDB &85.50&89.80&\textbf{93.80}&90.80&N/A&76.43&86.33&76.50&	86.91
\\
MR &76.70&75.50&73.30&77.00 &N/A&74.75&83.00&76.83&	\textbf{84.33}
 \\
\hline  
Avg &86.09&88.61&87.86&90.16&90.50&80.21&90.82&83.35&\textbf{91.49}
\\
\hline 
\end{tabular}
\caption{Results on FDU-MTL in the multi-domain setting. BERT-CE$^*$ and COBE$^*$ refer to the models fixing the parameters of BERT, and only tuning the parameters of MLP layer.}
\label{table3}
\end{table*}

\begin{table}[!htbp] \small
\centering
\vspace{1mm}
\begin{tabular}{l|cccc}   
\hline  
&ASP&DSR-at &DAEA&COBE  \\
\hline
\hline
Books &81.50 &85.80& 87.30 &\textbf{90.67}\\
Electronics&83.80&89.50 &85.80&\textbf{92.33 }\\
DVD &84.50&86.30 &88.80 &{87.50}
\\
Kitchen&87.50&88.30&88.00&\textbf{90.75}
\\
Apparel&85.30&85.80& 88.00&\textbf{91.16}
\\
Camera&85.30&88.80 &90.00&\textbf{91.67}\\
Health&86.00
&90.50 &91.00&\textbf{94.33}\\
Music &81.30
&84.80& 86.50 &\textbf{89.17}
\\
Toys&88.00
&90.30 &90.30&\textbf{92.33}
\\
Video &86.80&85.30
&\textbf{91.30}&88.50
\\
Baby&86.50
&84.80 &90.30 &\textbf{93.17}
\\
Magazines&87.00
&84.00 &88.50&\textbf{90.50}\\
Software&87.00
&90.80
&89.80&\textbf{90.82}
\\
Sports&87.00
&87.00 &90.50&\textbf{92.15}\\
IMDB&84.00
&83.30 &85.80&\textbf{86.58}\\
MR&72.00&
76.30& 75.50&\textbf{78.91}\\
\hline
Avg&84.59
&86.35 &87.96&\textbf{90.03}
\\
\hline
\end{tabular}
\caption{Results on FDU-MTL in the 15-1 setting.}
\label{table151}
\end{table}

\textbf{Implementation Details.}
We perform experiments using the official pre-trained BERT model provided by Huggingface\footnote{https://huggingface.co/}.  We train our model on 1 GPU (Nvidia GTX2080Ti) using the Adam optimizer \cite{kingma2014adam}. For the cross-domain Amazon dataset (FDU-MTL), the max sequence length for BERT is 256 (128), and the batch size $M$ is 8 (32). The max sequence lengths are set in such values for comparison with previous models. The initial learning rate is 2e-5 (1e-4) for BERT-unfixed (BERT-fixed) models, and each model is trained for 20 epochs. The hyper-parameters of temperatures $\tau$ is 0.05, $T$ is 5, and the number of nearest neighbors $k$ is 3 (without losing generality, we do not search for the best hyper-parameters through grid-search). Through the training of our model, no development set is applied to find the best checkpoints, but stop until the training step is reached. During the test procedure, we adopt FAISS IndexFlat Index \cite{johnson2018training} to accelerate the speed to find the $k$ nearest neighbors. We average the results with 3 different random seeds.

\subsection{Results}
\textbf{Results on the Cross-Domain Amazon Dataset.}
The results are shown in Table \ref{table2}. Overall, our model COBE achieves state-of-the-art performances with an accuracy of 90.39\% on average for the 12 cross-domain tasks. It achieves state-of-the-art performance in 9 of 12 tasks. The result is 2.14\% higher than that of BERT-CE (88.25\%), which indicates that our proposed contrastive learning method can be more effective and generalized than methods based on cross-entropy loss. COBE is also 1.83\% higher than BERT-adv (88.56\%), which implies that directly pushing the representations of different domains with the same (different) labels close (apart) results in a strong performance on the cross-domain sentiment classification.

DAAT uses the unlabeled data from the source domain and the target domain to continually train BERT to mix the information of the source domain and target domain. Then the training objective of cross-entropy and the domain discriminator are jointly optimized to obtain the sentiment classification model.  The average accuracy of our model is 0.27\% higher than that of DAAT, which uses additional data to continually train BERT to transfer knowledge in the source domain to the target domain. Although DAAT achieves great performance, it is more time-consuming and resource-wasting compared with solely using contrastive learning.  In the tasks of E$\to$ B, E $\to$ D, and K $\to$ D, the accuracies of our model are smaller compared with DAAT, and the possible reason can be that the source domains' data have less shared information with the target domains. But with unlabeled data for continual training, some domain-specific information is extracted in DAAT and further results in a better performance.

Moreover, the average accuracy of the model COBE$^*$ (82.05\%) outperforms that of BERT-CE$^*$ (54.86\%) with a large margin, where the parameters of BERT are fixed (corresponding to the scenario that pre-trained models are too large for fine-tuning). The model BERT-CE$^*$ fails to predict the sentiments of the target domain using cross-entropy-based methods, but with contrastive learning, it can obtain strong results (similar performance to BERT-CE). But the performance of models fixing BERT parameters is still largely worse than that of unfixed models. 

\begin{figure}[t]
    \centering
    \includegraphics[width = 0.95\hsize]{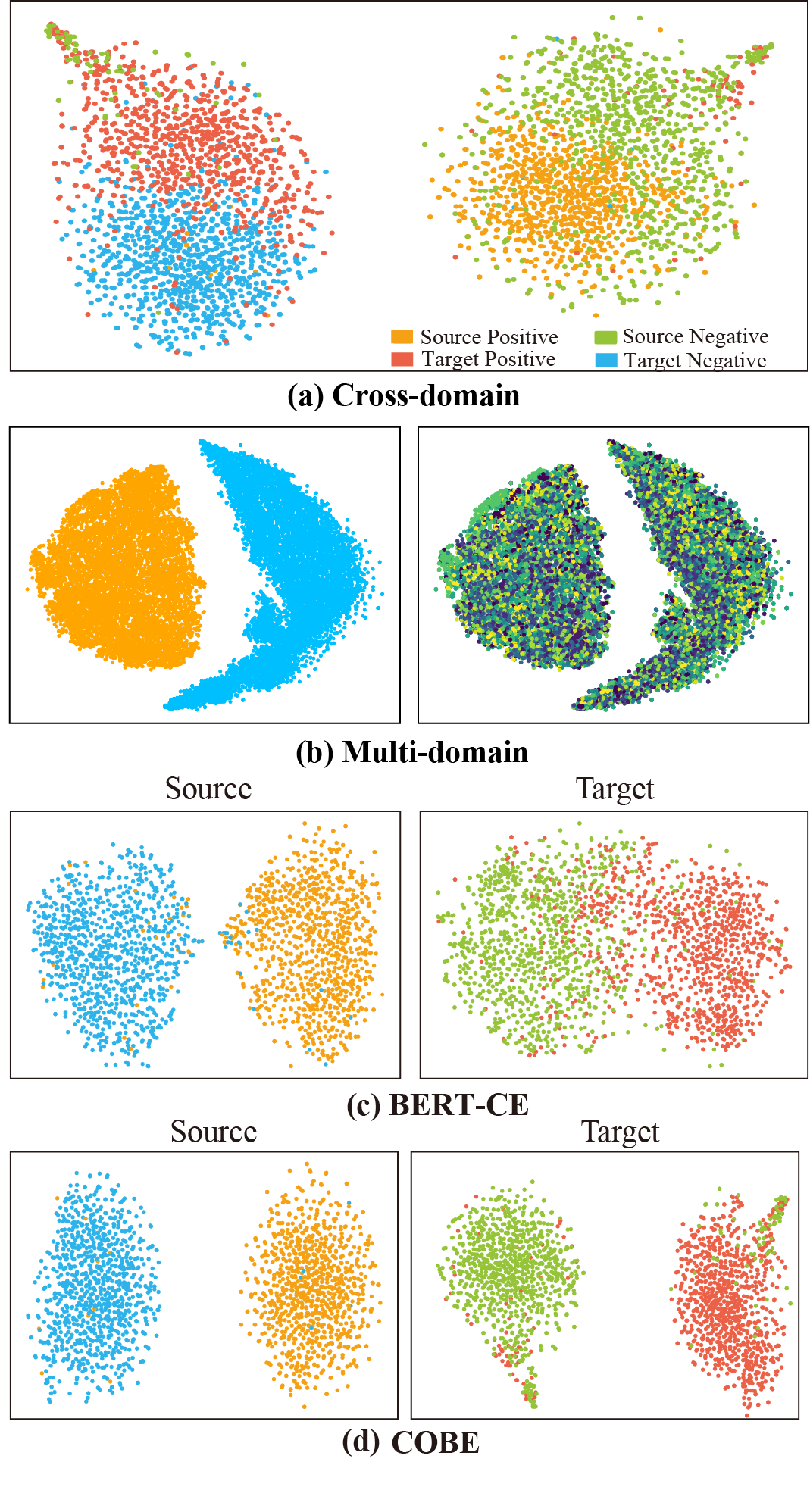}
    \caption{Visualizations of the sentence representations. We use t-SNE to transfer 192-dimensional feature space into two-dimensional space. (a)(c)(d) for the representations of the B->K task in the cross-domain Amazon dataset; (b) for those of the multi-domain task in FDU-MTL (colors in left for sentiment labels and right for domains).}
    \label{vis1}
\end{figure}
\textbf{Results on FDU-MTL.}  First, we test our model in the multi-domain setting, training the model on the data of 16 domains and evaluating it on the whole test data. The results are shown in Table \ref{table3}. Our model achieves the state-of-the-art performance with an accuracy of 91.49\%  on average, and in the 12 of 16 domains, it achieves the state-of-the-art performance. The accuracy is 0.67\% higher than that of BERT-CE, and 0.99\% higher than that of DAEA-B. In particular,  using BERT-CE solely in the multi-domain setting can achieve competitive performance (90.82\%), which is neglected by previous studies. The accuracy of our model COBE on the IMDB data is lower than DSA with a large margin, which may result from the max sequence length for BERT being 128, much smaller than the average sequence length in IMDB (128 to 256). Our model COBE$^*$ achieves an accuracy of 83.35\%  in the multi-domain setting, which is also higher than that of BERT-CE$^*$ with a margin of 3.14\%. 

Then we also evaluate our model in the 15-1 settings, referring to that we train the model on 15 domains and test it on the domain left (shown in Table \ref{table151}. Our model achieves state-of-the-art performance with an accuracy of 90.03\%. The accuracies in 14 of 16 tasks are larger than that in previous studies. It is 2.07\% higher than the average accuracy of DAEA. The experimental results also show that contrastive learning can perform better than cross-entropy-based models with adversarial training for cross-domain sentiment analysis.

 
  \begin{table*}[thp]\small
  \centering
\begin{tabular}{p{0.7\textwidth}p{0.1\textwidth}p{0.1\textwidth}}
\hline

{\textbf{Text}} & \textbf{Gold Label} & \textbf{Output} \\
\hline
\hline

\textbf{Test Data.}   This story is true to life living in south west and west phila. It brought back many memories and changing the names did not bother me. I really enjoyed reading about life the way it was back in the 55 to 70 era.& Positive & Positive \\

$k$ \textbf{nearest neighbors.} \\ 
(1) Have to be honest and say that I haven't seen many independent films, but I thought this one was very well done.  The direction and cinematography were engaging without becoming a distraction. & Positive\\
(2) I bought this wireless weather station as a gift. The recipient loves it. For the price, he is really enjoying it. & Positive\\
(3) I think j-14 is a really good magazine if u like to hear the latest gossip about all your favourite celebrity 's, or if u like to get nice posters of all the hot celebrity 's. & Positive\\
\hline
\end{tabular}
\caption{Case Study on FDU-MTL.}
\label{case}
\end{table*}

\subsection{Visualization}
The visualization of the sentence representations $\textbf{h}_i$ in COBE is shown in Figure \ref{vis1}. For the B->K (Books->Kitchen) task in Figure \ref{vis1} (a), first, the representations of positive and negative data are separated acutely with a large margin between each other. Second, representations of source and target domains with the same labels are close to each other, which means the knowledge learned from the source domain is transferred to the target domain effectively. 
For the multi-domain setting in Figure \ref{vis1} (b) (left), we can observe that the representations with the same labels are separated into different clusters w.r.t the labels, and in the sentence representations with the same label but different domains mix up well, which satisfy the requirements of the cross-domain sentiment analysis. 

To further compare the contrastive learning method with cross-entropy-based methods, we illustrate the representations of the source domain and the target domain in COBE, BERT-CE in Figure \ref{vis1} (c)(d), respectively (visualizations of COBE$^*$ and BERT-CE$^*$ are shown in Appendix). Obviously, the sentence representations are separated  in the target domain in BERT-CE less effective than that in COBE.  The visualizations show the effectiveness of contrastive learning in transferring the learned knowledge in the source domain to the target domain. Meanwhile, it demonstrates operating the sentence representations in the feature space has a strong generalization ability in the cross-domain sentiment analysis tasks.



\subsection{Robustness Analysis}
We evaluate our model on adversarial samples generated by using the well-known
substitution-based adversarial attack method--Textfooler \cite{jin2020bert}. Given an input $X_i$ and a pre-trained classification model $F$, a valid adversarial sample $X_i^{adv}$ should conform the following requirements:
\begin{equation}
    F(X_i) \neq F(X_i^{adv}), \  \ Sim(X_i,X_i^{adv})\ge \epsilon.
\end{equation}
where $Sim$ is a similarity function and $\epsilon$ is the minimum similarity between the original input and the adversarial sample, which is often a semantic and syntactic similarity function. The details for generation refer to \citet{jin2020bert}. An  adversarial sample is shown in Table \ref{ad}, where the sentence semantic information is not corrupted, but some words are replaced.

We test our model (trained using Books data in the cross-domain Amazon dataset) with 200 adversarial samples from the Kitchen domain, and our model (trained using multi-domain data in FDU-MTL) with 200 adversarial samples randomly selected from the multi-domain test data.  The results are shown in Table \ref{robust}. Our model COBE achieves 78.00\%  and 81.00\% accuracies for the two kind of adversarial data, which are 6.5\% and 7.5\% higher than BERT-CE. Meanwhile, the model COBE$^*$ outperforms BERT-CE$^*$ with a large margin (26.5\% and 23\%). The results demonstrate that contrastive-learning based models have better robustness than cross-entropy-based models.

\begin{table}[t]\small
    \centering
    \begin{tabular}{p{0.45\textwidth}}
        \hline
\textbf{Original Text}: DEF. NOT A GOOD TANK. You look at them in a picture frame, the fish are crammed in there.  \\
\hline
\textbf{Adversarial Text}: DEF. Not a \textcolor{red}{alright} tank you look at them in a \textcolor{red}{photography sashes} the fish are \textcolor{red}{teeming} in there.  \\
\hline
    \end{tabular}
    \caption{Adversarial example based on TextFooler.}
    \label{ad}
\end{table}

\begin{table}[t] \small
\centering
\vspace{1mm}
\begin{tabular}{l|cccc}   
\hline  
&BERT-CE$^*$&BERT-CE &COBE$^*$&COBE  \\
\hline
\hline
Books&42.50 &71.50&69.00&\textbf{78.00}\\ 
Multi-&49.50 &73.50&72.50&\textbf{81.00}\\
\hline
\end{tabular}
\caption{Results on the adversarial samples. Books for the trained model using the data of Books domain in the cross-domain Amazon dataset, and Multi- for the trained model using multi-domain data in FDU-MTL. }
\label{robust}
\end{table}

\subsection{Case Study}
The case study is shown in Table \ref{case}. As can be  observed, the $k$ nearest neighbors of the test data (Books) are reviews from different domains (Video, Electronics and Magazines) with positive labels, and it outputs the correct label for the test data. Note that the key sentiment information is similar for the original text and neighbors in the case such as `enjoy', `engaging', `enjoying' and `favorite'.  It shows that our model can learn effective information from multi-domain data for the sentiment classification task, and the representations of different domains mix up well, which serve as a strong sentiment knowledge base for the classification.

\section{Conclusion}
We explored the contrastive learning method in the cross-domain sentiment analysis task. We proposed a suitable contrastive loss for the supervised sentiment analysis task with the in-batch negatives method. Experiments on two standard datasets showed the effectiveness of our model. Visualizations also demonstrated the effectiveness of transferring knowledge learned in the source domain to the target domain. We also showed that our model has stronger robustness than cross-entropy-based models through the adversarial test.

\section{Ethical Statement}
We honor the ACL Code of Ethics. No private data or non-public information was used in this work.

\bibliography{anthology}
\clearpage
\appendix
\section{Statistics for FDU-MTL.}
\begin{table}[!ph]\small
\centering

\vspace{1mm}
\begin{tabular}{l|cccc}   
\hline  
Domain& Train &Dev&Test &Avg. Length \\
\hline
\hline
Books &1400&200&400&159
\\
Electronics&1398&200&400&101
\\
DVD &1400&200&400&173
\\
Kitchen&1400&200&400&89
\\
Apparel&1400&200&400&57
\\
Camera&1397&200&400&130
\\
Health&1400&200&400&81
\\
Music &1400&200&400&136
\\
Toys&1400&200&400&90
\\
Video &1400&200&400&156
\\
Baby&1300&200&400&104
\\
Magazines&1370&200&400&117
\\
Software&1315&200&400&129
\\
Sports&1400&200&400&94
\\
IMDB&1400&200&400&269
\\
MR&1400&200&400&21
\\
\hline
\end{tabular}
\caption{Statistics of FDU-MTL.}
\label{table1}
\end{table}

\section{Reconstruction Loss}
We attempt to reconstruct the  representations of BERT which means another MLP layer is applied by $\textbf{h}_i^{rec} = MLP(\textbf{{h}}_i)$. Then a reconstruction loss of MSE (mean-squared loss) is added to retain the semantic information, $\mathcal{L}_{rec} = ||\textbf{h}_i^{rec} - \textbf{h}_i^{CLS}||$ as \cite{zhao2021relation}. But little improvement (an average accuracy  of 90.81\% on the FDU-MTL multi-domain setting and 90.13\% on the cross-domain Amazon dataset) is obtained, which is 0.68\% and 0.26\% lower than COBE, respectively. 
It indicates that the reconstruction loss is not suitable for the task of cross-domain sentiment analysis.

\section{SCL Loss}
To verify the effectiveness of our propose loss function, we compare our contrastive learning loss with the SCL loss \cite{gunel2020supervised}, which can be formulated as follows:
\begin{equation}\small
    \mathcal{L}_{SCL} = - \sum_i^M \frac{1}{|N_i^+|} \sum_{k\in N_i^+} log \frac{ exp(sim(\textbf{h}_i, \textbf{h}_k)/\tau)}{\sum_{i\neq j}^M exp(sim(\textbf{h}_i,\textbf{h}_j)/\tau)}
\end{equation}

In fairness, we use the kNN predictor the same as our proposed model. The model with SCL loss achieves an average accuracy of 91.03\% on the FDU-MTL multi-domain setting and 90.05\% on the cross-domain Amazon dataset (0.46\% and 0.34\% lower than COBE, respectively). The experiments prove the effectiveness of our proposed loss function with the in-batch negative samples, which aims to tighten all the samples of the same labels as positive pairs. The conclusion is different from that in \citet{khosla2020supervised}, whose experiments demonstrate that the separately calculating each positive pair separately (SCL loss) achieves better results for image classification. It may results from the reason that the batch size influences the results of the two methods, and our batch sizes (8 and 32) are comparatively smaller compared with their study (6144), which may motivate further theoretical analysis.

\section{Influence of $k$}    
\begin{figure}
    \centering
    \includegraphics[width = 0.9\hsize]{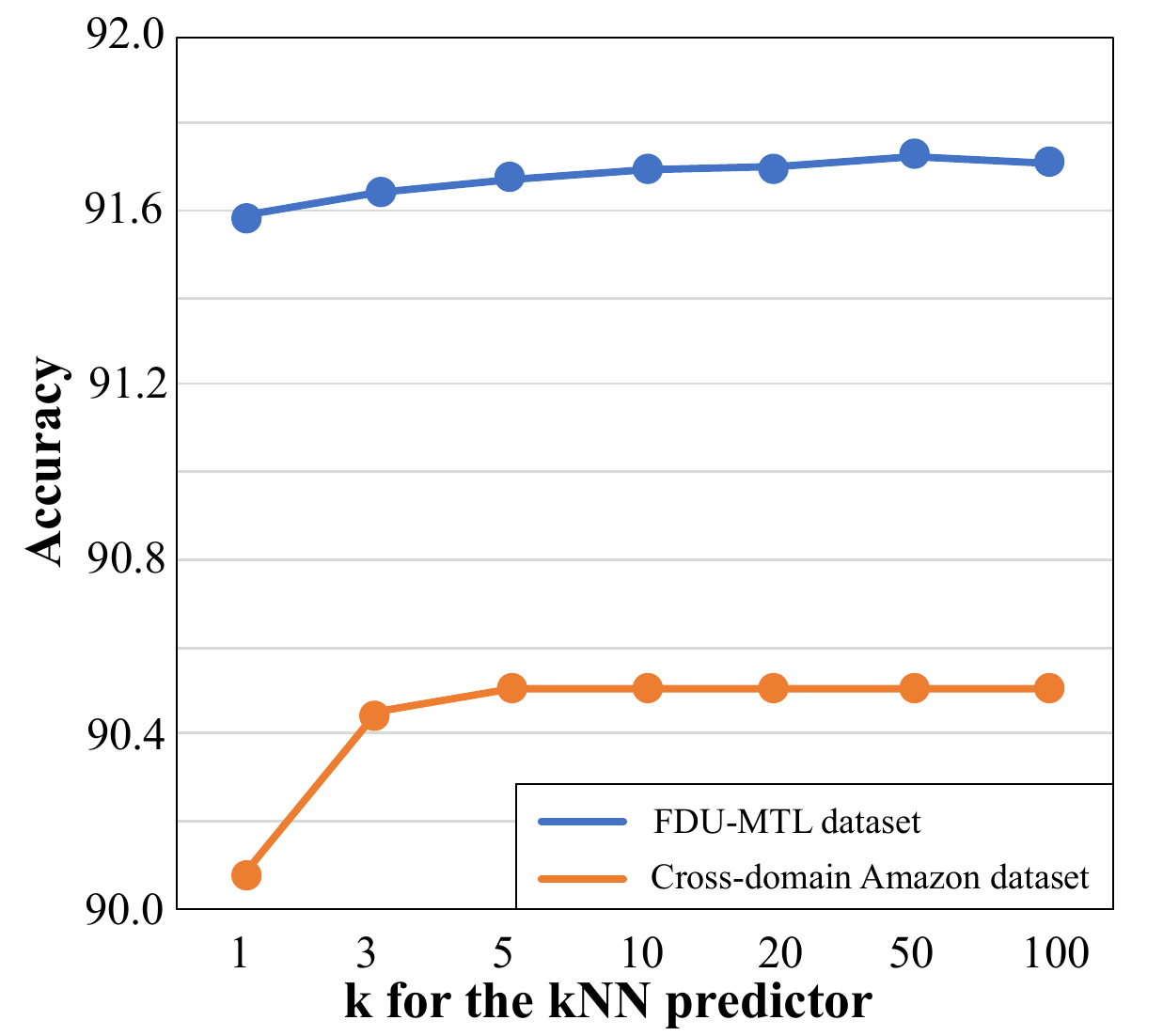}
    \caption{Evaluation with respect to different numbers of $k$ for one random seed.}
    \label{k}
\end{figure}

In order to discover the sensitivity of our model to the influence of k for the kNN predictor (shown in Figure \ref{k}), we evaluate our model with respect to the different numbers of $k$. As observed, the accuracies of COBE stop to increase and keep stable when $k>=5$, which indicates that the model is little sensitive with the hyper-parameter $k$. The phenomenon demonstrates that the sentence representations learned by COBE are effectively separated and stable for classification.

 \begin{table*}[thp]\small
  \centering
\begin{tabular}{p{0.5\textwidth}p{0.5\textwidth}}
\hline

{\textbf{Original Text}} & \textbf{Adversarial Text} \\
\hline
\hline
Very nice iron!.	This is a great iron.  It's quite heavy, but I like that.  It really gets out the wrinkles.  I don't even mind ironing any more.& Awfully sweet iron! This is a whopping iron it 's quite heavy, but I like that it really gets out the wrinkles I don't even mind ironing any more.  \\ \hline
A good idea, disappointing in use.	These silicone pot holders are indeed brightly colored, easy to wash in the dishwasher, and protective even when wet.  They are also clumsily stiff at the same time as they are slippery, the net result being a miserable failure in the kitchen.  They are useful for protecting a counter from a hot pot, but not for picking the hot pot up. &
A good ideas, agonizing in use these silicon pot holders are indeed brightly colour, easy to wash in the dishwasher, and protective even when clammy they are also clumsily painstaking at the same time as they are slippery, the net raison being a miserable failure in the kitchen they are useful for protecting a counter from a hot pot, but not for picking the hot pot up.\\

\hline
Love this piece.	I just bought this piece and tried it out. I love the size and no drip mouth.The color is beautiful and its so pretty on my buffet. & Like this pieces I just obtained this pieces and attempts it out. I luv the size and no drip mouths the colorful is beautiful and its however rather on my buffet.
\\
\hline
\end{tabular}
\caption{Adversarial Samples.} 
\label{adv}
\end{table*}

\begin{figure*}[h]
    \centering
    \includegraphics[width = \hsize]{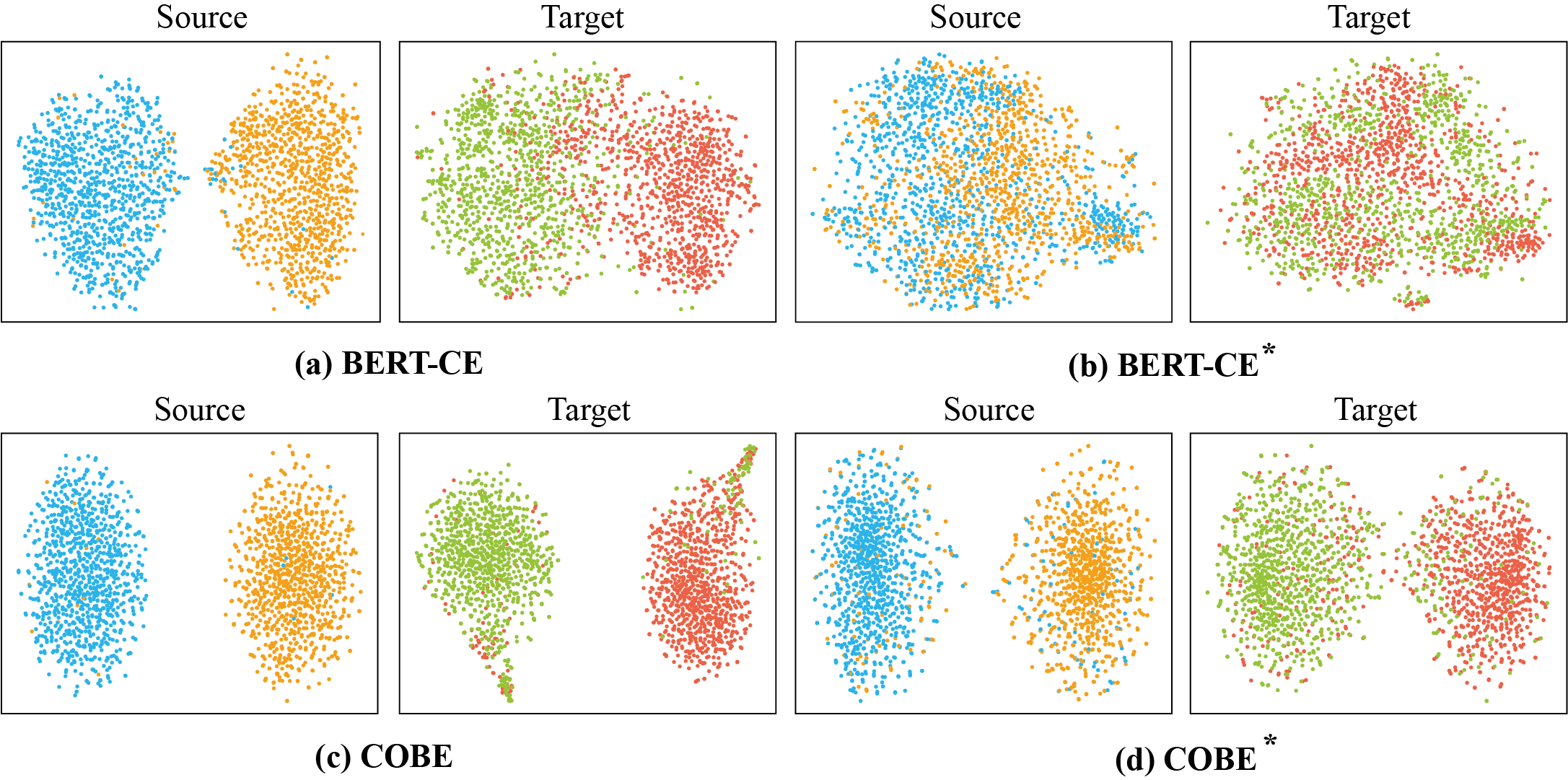}
    \caption{Visualization of sentence representation obtained from BERT and COBE. We use t-SNE to transfer the feature space into two-dimensional space for the  B$\to$K task.}
    \label{vis222}
\end{figure*}

\end{document}